\documentclass[runningheads]{llncs}

\usepackage{eccv}

\usepackage{eccvabbrv}

\usepackage{graphicx}
\usepackage{booktabs}

\usepackage{epsfig}
\usepackage{arydshln}
\usepackage{subcaption}

\usepackage[accsupp]{axessibility}  % Improves PDF readability for those with disabilities.

\usepackage{hyperref}

\usepackage{orcidlink}

\begin{document}

\title{SOOD-ImageNet: a Large-Scale Dataset for Semantic Out-Of-Distribution Image Classification and Semantic Segmentation} 

\titlerunning{A Semantic Out-Of-Distribution Dataset for Computer Vision}

\author{Alberto Bacchin\orcidlink{0000-0002-2945-8758} \and
Davide Allegro\orcidlink{0009-0008-1180-9290} \and
Stefano Ghidoni\orcidlink{0000-0003-3406-8719} \and
Emanuele Menegatti\orcidlink{0000-0001-5794-9979}
}

\authorrunning{A.~Bacchin et al.}

\institute{Department of Information Engineering, University of Padova, Padova, Italy\\
\email{\{bacchinalb,allegrodav,ghidoni,emg\}@dei.unipd.it}}

\maketitle

\begin{abstract}
Out-of-Distribution (OOD) detection in computer vision is a crucial research area, with related benchmarks playing a vital role in assessing the generalizability of models and their applicability in real-world scenarios. However, existing OOD benchmarks in the literature suffer from two main limitations: (1) they often overlook semantic shift as a potential challenge, and (2) their scale is limited compared to the large datasets used to train modern models. To address these gaps, we introduce SOOD-ImageNet, a novel dataset comprising around 1.6M images across 56 classes, designed for common computer vision tasks such as image classification and semantic segmentation under OOD conditions, with a particular focus on the issue of semantic shift. We ensured the necessary scalability and quality by developing an innovative data engine that leverages the capabilities of modern vision-language models, complemented by accurate human checks. Through extensive training and evaluation of various models on SOOD-ImageNet, we showcase its potential to significantly advance OOD research in computer vision. The project page is available at \texttt{\url{https://github.com/bach05/SOODImageNet.git}}.

\end{abstract}

\section{Introduction}
\label{sec:intro}

The research on computer vision is currently driven by Deep Learning (DL) methods, which are typically trained and benchmarked using public datasets, such as ImageNet~\cite{imagenet_cvpr09}, Pascal VOC~\cite{pascal} and CIFAR-100~\cite{krizhevsky2009learning}. Standard benchmarks in computer vision assume that both train and test data are similarly distributed (IID), \ie train and test sets are just a random split of the same dataset. 
%However, a drop in performance is observed when the models are deployed in a context where the data follows a different distribution~\cite{ood-cv}. This problem, known as Out-Of-Distribution (OOD) generalization, is crucial for comparing and deploying DL models in real-world scenarios, and it is a significant research topic today~\cite{zhang2024openood}. 
However, a drop in performance is observed when models are deployed in environments where the data distribution differs from that of the training set~\cite{ood-cv}. This challenge, referred to as Out-Of-Distribution (OOD) generalization, is crucial for effectively comparing and deploying DL models in real-world scenarios, making it a significant focus of current research on computer vision~\cite{zhang2024openood}.
To address OOD challenges, it is essential to have appropriate datasets for training and benchmarking DL models under OOD conditions. Consequently, new datasets and benchmarks have been proposed. For instance, the OOD-CV dataset by Zhao~\etal~\cite{ood-cv} represents OOD samples as variations in pose, shape, texture, context, and weather conditions relative to the training set. Another relevant approach is the Semantic Shift Benchmark (SSB)~\cite{vaze2022openset-semanticshift}, which targets Open Set Recognition (OSR)~\cite{open-set-scheirer}. OSR involves identifying input examples that do not belong to any of the known classes (i.e., they exhibit a semantic shift) and correctly classifying them as "unknown" or "out-of-distribution". A good example is a face recognition system, which must correctly recognize only the known faces while identifying any unfamiliar ones as unknown. 
\begin{figure}[tb]
  \centering
    \includegraphics[width=\linewidth]{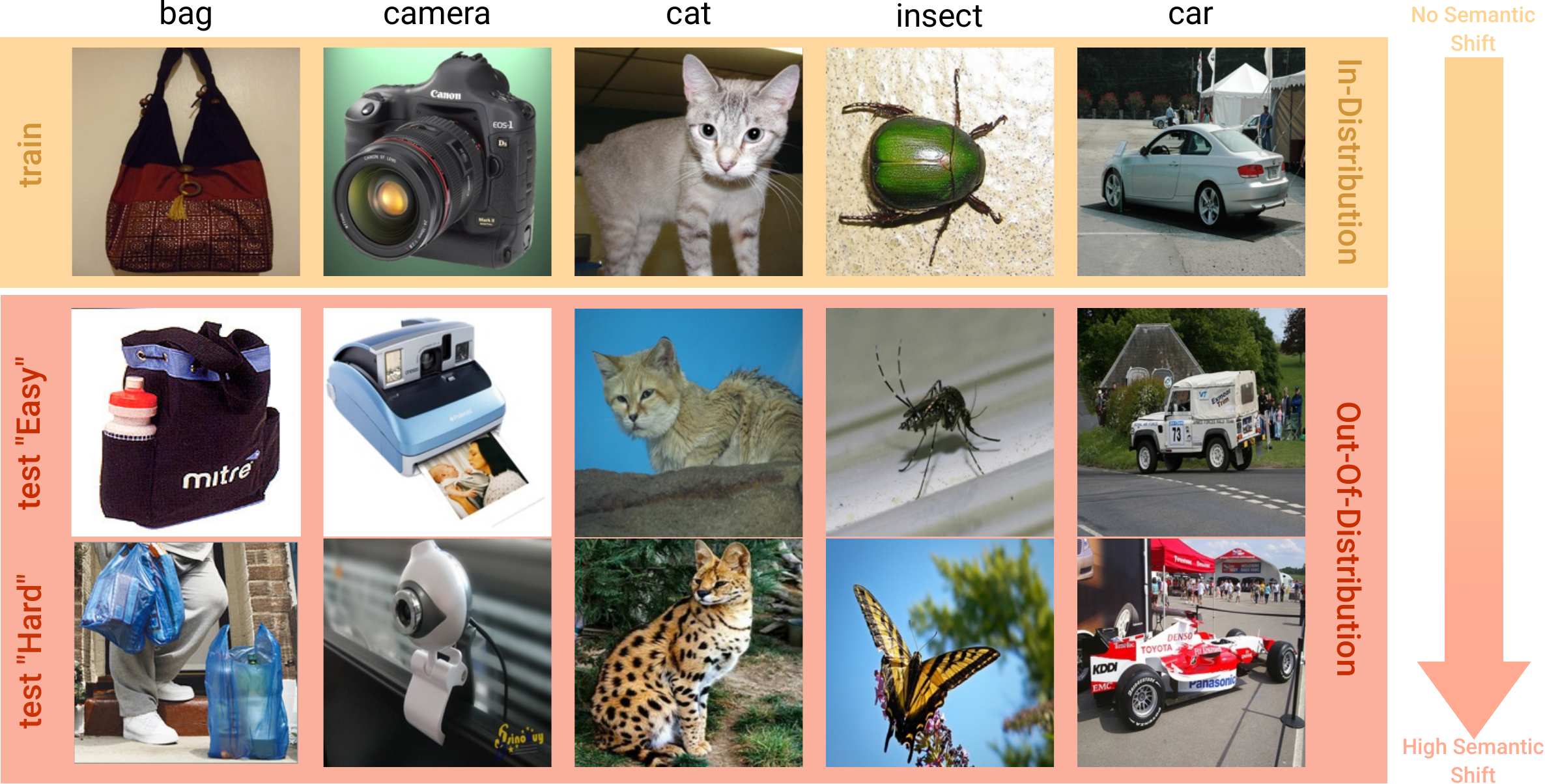}
    \caption{Example of images taken from the proposed SOOD-ImageNet. It can be noted the increasing semantic shift from the train to the test "Hard" data.}
  \label{fig:test_samples1}
\end{figure}
These two settings highlight a gap: the former involves the same train and test classes with contextual variations, without semantic shift (covariated shift~\cite{zhang2024openood}), while the latter aims to reject classes which presents a semantic shift. However, these benchmarks do not explore models' abilities to correctly classify -- within training categories -- examples that present semantic shifts, such as shown in Figure~\ref{fig:test_samples1}. 
%However, these benchmarks do not explore models' abilities to handle semantic shifts within the training classes. 
For example, if a model is trained on car and chair classes, can it correctly classify a wheelchair in the chair class? This capability is relevant in various fields, such as autonomous driving~\cite{filos2020can}, agriculture~\cite{Angarano_2024_CVPR_agriculture} and waste management~\cite{pmlr-v220-bashkirova23a_visda}. We refer to this problem as Semantic Out-Of-Distribution (SOOD) shift. Despite some of the datasets already proposed in the literature could be used to address this task, they presents some limitations both like the limited number of examples and classes~\cite{vim_openimage-o}, or their scope, i.e., focusing on specific domains such as natural species~\cite{iNaturalist} or textures~\cite{texture}. With the rise of large DL models, large-scale datasets are required for successful training and a broad coverage of different categories and concepts is important to assess the generalization performance of the models. Moreover, previously proposed datasets primarily cover classification and object detection tasks, neglecting the semantic segmentation task, which has historically been very relevant in computer vision and remains a flourishing research area~\cite{semantic_segmentation_survey}. To tackle the aforementioned issues, we introduce SOOD-ImageNet, a large-scale dataset designed to test the semantic Out-Of-Distribution performance of DL models on both classification and semantic segmentation tasks. 
\begin{figure}[tb]
  \centering
  \begin{subfigure}{0.95\linewidth}
  \includegraphics[width=\linewidth]{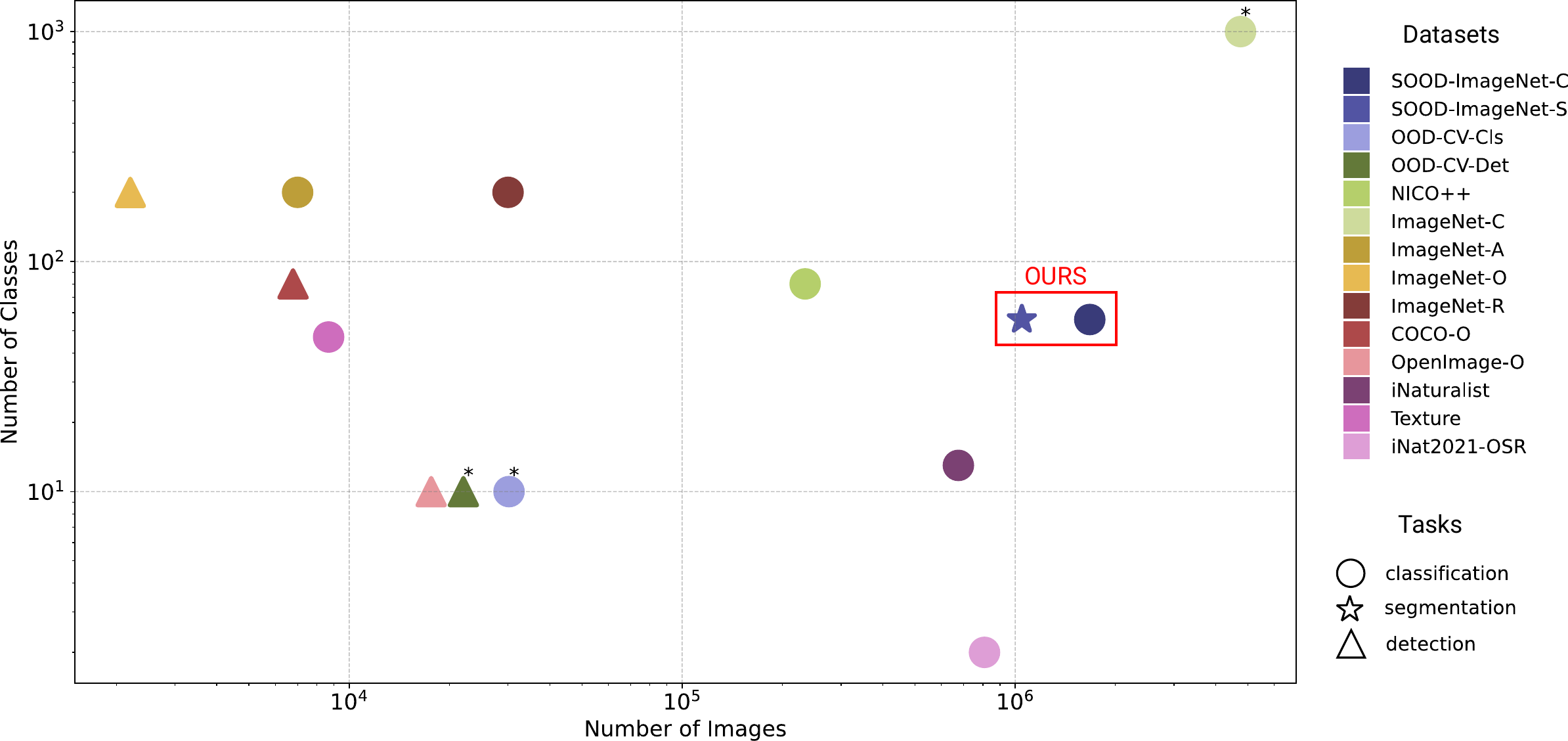}
  \caption{Number of classes and number images for each dataset. (*) indicates presence of synthetic data. To the best of our knowledge, SOOD-ImageNet is the largest non-synthetic dataset available, and it uniquely offers additional labels for semantic segmentation.}
  \label{fig:dataset_sizes_comparison}
  \end{subfigure}
  \vspace{0.5cm}
  \begin{subfigure}{0.95\linewidth}
    \includegraphics[width=\linewidth]{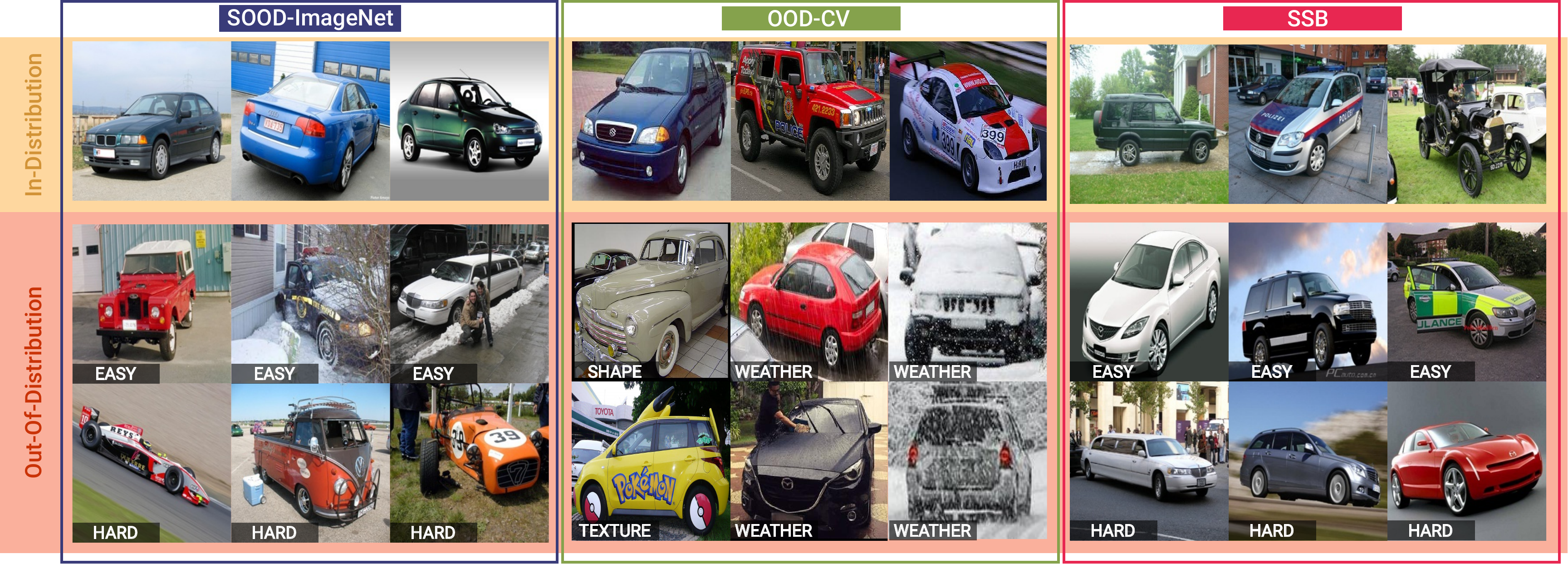}
    \caption{Examples of the "car" class from different datasets: SOOD-ImageNet (blue), OOD-CV (green), and SSB (red).  Black labels indicate the nuisance or entity of the shift. Notably, SOOD-ImageNet demonstrates a better representation of semantic shift, while the other datasets primarily focus on different types of shifts.}
    \label{fig:dataset_image_comparison}
  \end{subfigure}
  \caption{Comparison of different datasets for OOD in computer vision}
  \label{fig:augmentations}
\end{figure}
To achieve both scalability and quality, we developed a new data engine that combines automatic and manual procedures. While manual checks ensure high-quality data~\cite{vim_openimage-o}, automatic labeling allows us to drastically scale up the size of the dataset. The proposed data engine applies various filtering and re-labeling steps to ImageNet-21K-P~\cite{imagenet21k-p}. As illustrated in Figure~\ref{fig:SOOD_imagenet_fw}, it begins with an initial filtering step based on natural language relationships between the classes in ImageNet-21K-P. Subsequently, it leverages Vision-Language Models (VLMs) to incorporate visual cues alongside language, further refining and scoring the data. These scores are ultimately used to define IID and OOD samples. While previous works~\cite{imagenet-a, iNatural2021-osr} have proposed scoring strategies with neural networks, they focused solely on image embeddings. By integrating language cues, we introduce stronger priors into the proposed method to get more reliable scores~\cite{Conde_2021_CVPR_clip_art, Khandelwal_2022_CVPR_clip_embodAI}. 
%
% \begin{figure}[tb]
%   \centering
%   \includegraphics[width=\linewidth]{imgs/datasets_sizes.eps}
%   \caption{Comparison of different datasets for OOD in computer vision, in terms of number of classes and images. (*) indicates presence of synthetic data. To the best of our knowledge, SOOD-ImageNet is the largest, non-synthetic dataset in literature, providing additional labels for semantic segmentation.}
%   \label{fig:dataset_sizes_comparison}
% \end{figure}
%
%
This strategy significantly reduced the effort required for human annotators compared to fully manual annotations, employing approximately 12 hours of human labor in the entire process. As a result, we created SOOD-ImageNet-C for image classification, containing 1.6M images and 56 classes. The dataset is strategically divided based on the scores: around 1M images for IID training and approximately 0.6M images for OOD tests. The test set was further partitioned into "Easy" and "Hard" splits, each reflecting an increasing degree of semantic shift. This allows for a thorough evaluation of the model's ability to generalize beyond the training distribution. To obtain  SOOD-ImageNet-S for semantic segmentation, we employed VLMs to automatically label both the training and test sets. The test set labels were then meticulously verified by human annotators, resulting in 8335 high-quality segmentation masks for OOD tests. As shown in Fig.~\ref{fig:dataset_sizes_comparison}, our SOOD-ImageNet is the largest non-synthetic dataset for computer vision tasks in the context of OOD research, also providing significant variability of classes. Additionally, our data engine offers more flexibility than a fully manual pipeline, as tuning the hyperparameters of the scoring function allows for the creation of different splits with varying semantic shifts and granularity. We conducted extensive experiments to showcase the challenges introduced by the SOOD-ImageNet dataset. These experiments included comparisons of various model architectures and the application of state-of-the-art data augmentation techniques. Despite these efforts, we found that DL models consistently struggled with the challenging conditions presented by the SOOD generalization problem. To ensure a comprehensive evaluation, we also tested large pre-trained models, which produced similar results. These findings highlight that the issue of semantic shift remains a significant and unresolved problem in the field. We believe that our data engine makes significant contributions to OOD research in computer vision by providing (1) a large-scale dataset for image classification, (2) the first dataset specifically designed for semantic segmentation, and (3) a robust benchmark for assessing semantic shifts, as shown in Figure~\ref{fig:dataset_image_comparison}. 
\section{Related Works}
\label{sec:related_works}
\subsubsection{Out-Of-Distribution Generalization Benchmarks.} Out-Of-Distribution generalization involves testing model performance under distributional shifts with respect to the training data. As pointed out by Ye~\etal~\cite{Ye_2022_CVPR_ood_bench}, this task can be approached in various ways. NICO++~\cite{nico++} addresses it as a problem of Domain Generalization by introducing a dataset where each image is labeled with both a categorical label describing the main subject and a domain label describing the context. For example, an image labeled as [dog, grass] is expected to depict a dog on a lawn, while [dog, indoor] indicates a dog inside a house. Following the same philosophy, Mao~\etal\cite{coco-o} proposed COCO-O for OOD object detection, featuring 6 different domain variations. However, contextual domain is not the only relevant distribution shift. ImageNet-R~\cite{imagenet-r} explores the influence of style in the images, introducing several renditions of ImageNet-1K classes such as "cartoon", "painting", "sculpture" and more. Similarly, OOD-CV~\cite{ood-cv} introduce variations in pose, shape, texture, context, and weather conditions starting from PASCAL3D+~\cite{pascal3d+}, providing a benchmark with broader real-world applicability. These datasets share a common collection procedure: they start with categories from another public dataset and collect variations from the internet using suitable queries, such as "car+snow," followed by manual checks to remove outliers. This approach does not ensure that all samples are OOD, as it cannot guarantee that the training set does not contain the selected variations. A possible solution is to introduce synthetic modifications (\eg blur, Gaussian noise, etc...) to the images to generate OOD samples, as in ImageNet-C~\cite{imagenet-c}. While generating large amounts of synthetic samples is relatively easy, such data may not accurately capture real-world conditions. On the other hand, we take a different approach to address the issue. Our data engine starts with a massive dataset and partitions it into IID and OOD subsets, with no overlap. This method guarantees that OOD samples are distinct from the training data to provide a reliable assessment of model performance under distributional shifts. 
\subsubsection{Open Set Recognition Benchmarks.} Open Set Recognition (OSR), also known as novelty detection, was initially introduced by Scheirer \etal~\cite{open-set-scheirer}. OSR involves recognizing objects from unknown classes at test time, i.e., classes not present in the training data. In contrast to the closed-set setting, where new instances of the same classes are presented at test time, OSR requires the model to predict an additional "unknown" class when encountering samples from novel classes. This intrinsically models an OOD setting. However, unlike the previously mentioned works that focus on covariate shift\cite{zhang2024openood}, OSR emphasizes semantic shift, meaning that OOD samples come from different classes. The Semantic Shift Benchmark (SSB)~\cite{vaze2022openset-semanticshift} is a prime example of an OSR benchmark. The authors leveraged the fact that ImageNet-21K-P is a strict superset of ImageNet-1K, allowing them to sample disjoint novel category sets from the former. Novel categories present a semantic shift with respect to categories in ImageNet-1K and are organized into "Easy" and "Hard" sets, based on the degree of shift measured as the path distance in the semantic tree of ImageNet-21K-P. The hypothesis is that larger shifts make it easier for the sample to be recognized as "unknown". Another example is iNat2021-OSR, proposed by Lang et al.\cite{iNatural2021-osr}. Building on iNat2021\cite{inat}, a visual dataset of natural species, the authors used the hops in the taxonomy to measure the semantic shift between categories. With up to seven hop distances for the two super-categories of "Birds" and "Insects", this benchmark allows for a finer-grained measurement of OSR performance. Despite its large-scale size in terms of number of images, iNat2021-OSR is focused only on natural species, with limited impact in other domains. Texture~\cite{texture} is another dataset often used for benchmarking in OSR~\cite{mos} since it contains images of textures that are not commonly present in other datasets. However, it does not cover realistic scenarios, limiting its usefulness for practical applications. A more balanced benchmark is proposed by Wang~\etal~\cite{vim_openimage-o} with OpenImage-O. The dataset includes high-quality, manually selected samples from the test set of OpenImage~\cite{kuznetsova2020openimages}. Human annotators were asked to determine if an example was OOD, supported by the category labels and most similar images to the test sample in each category, measured by cosine similarity in a feature space. The purely manual labeling procedure has two main drawbacks: (1) the scale of the dataset remains limited and (2) the concept of "OOD-ness" may vary from person to person and lacks fine-grained evaluation. With SOOD-ImageNet, we aim to overcome these limitations by providing a large-scale dataset that covers a wide range of real-world categories and enables fine-grained evaluation.
\subsubsection{Semantic Out-Of-Distribution Generalization Benchmarks.} OSR is particularly relevant for applications like face recognition. However, in other domains, it is crucial to evaluate the generalization capabilities to novel classes that present a semantic shift from the training classes. For instance, in agriculture, weed detection is a significant challenge due to the existence of thousands of weed species and the lack of exhaustive training datasets~\cite{weed}. We have identified this capability as SOOD generalization. Although this concept is not yet prominently highlighted in the literature, there are strong connections to previous works that can be drawn. Thanks to its hierarchical organization, iNaturalist~\cite{iNaturalist} could be used for to test SOOD generalization. However, it is again strictly focused on natural species. ImageNet-A and ImageNet-O~\cite{imagenet-a} have been proposed to test SOOD generalization in extreme cases. Authors used a pre-trained ResNet-50~\cite{resnet} to select a set of adversarial examples from ImageNet-21K-P, creating an OOD benchmark for models trained on ImageNet-1K. Despite the great theoretical contributions, these datasets have limited applicability in real-world since the samples often represents rare conditions, \eg a false-color image of the Sun. SOOD-ImageNet contains a significant number of categories, with different grains of difficulty in order to gracefully test models in a more realistic scenarios. 
\section{Method}
\label{sec:method}
\begin{figure}[tb]
  \centering
    \includegraphics[width=\linewidth]{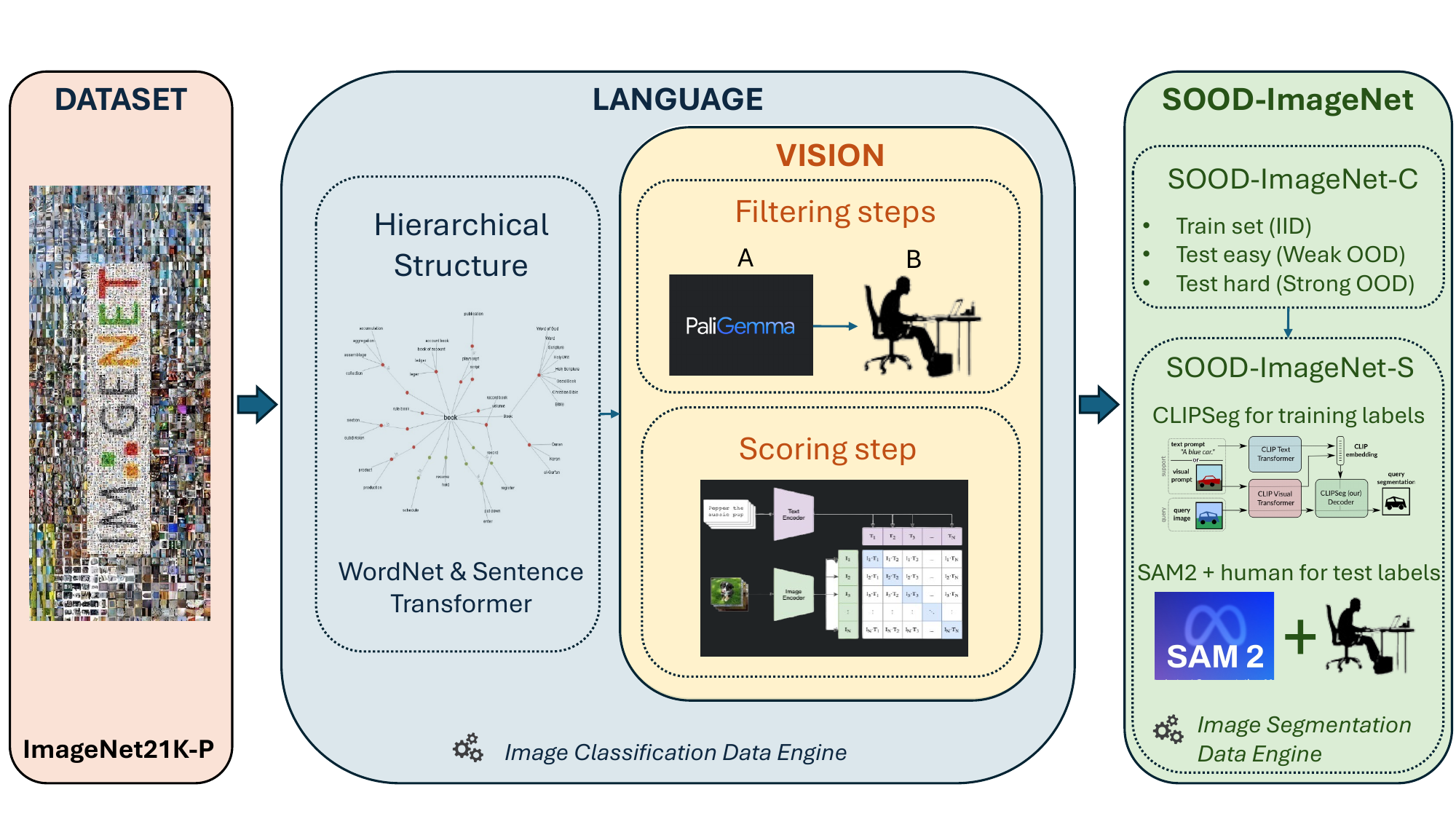}
    \caption{Pipeline of the SOOD-ImageNet dataset creation. Our data engine starts from ImageNet-21K-P and create a hierarchical structure using the semantics of language. Then VLMs are applied to filter, relabel and score the data, alongside with human checks.}
  \label{fig:SOOD_imagenet_fw}
\end{figure}

The proposed SOOD-ImageNet datasets aims to provide large-scale and high-quality benchmark for SOOD generalization of DL models. A more formal definition of SOOD generalization is as follow. Given a set of super-classes $\mathbb{C}$, for each $\mathcal{C} \in \mathbb{C}$, we define its sub-classes $\mathcal{S}_{\mathcal{C}} = \{ s_{\mathcal{C}} \mid s_{\mathcal{C}} \subset \mathcal{C} \}$. Each $s_{\mathcal{C}}$ defines a different semantic shift inside the super-class $S$. Given a scoring function $p : \mathcal{S}_{\mathcal{C}} \rightarrow [0, 1]$ that measures the correlation between the super-class $\mathcal{C}$ and its sub-class $s_{\mathcal{C}}$. Note that, up to a normalization factor, $p$ defines a probability mass function over $\mathcal{S}_{\mathcal{C}}$. It can be seen as the probability of the sub-class $s_{\mathcal{C}}$ to be a good representation of the super-class $\mathcal{C}$. We can partition $\mathcal{C}$ in $k$ subsets using the function $p$, such that $\mathcal{C}_{IID} = \{ s_{\mathcal{C}} \mid p(s_{\mathcal{C}}) > t^0 \}$ are the most representative sub-classes for $\mathcal{C}$, while $\mathcal{C}_{OOD}^i = \{ s_{\mathcal{C}} \mid t^i \le p(s_{\mathcal{C}}) < t^{i-1} \}$ with $i = 1 \, \dots \, k-1$ defines different grades semantic shifts with respect to $\mathcal{C}_{IID}$ ---where $t^i$ is a threshold. Performing the procedure for each $\mathcal{C} \in \mathbb{C}$ and gathering the splits, we obtain $\mathbb{C}_{iid}$ and $\mathbb{C}_{ood}^i$, with $i = 1 \, \dots \, k-1$. Given a model $M$, trained on $\mathbb{C}_{iid}$, the SOOD generalization benchmark aims to test the performance of $M$ on each $\mathbb{C}_{ood}^i$. To provide a SOOD generalization benchmark for classification (SOOD-ImageNet-C) and semantic segmentation (SOOD-ImageNet-S), we designed two complementary data engines, as illustrated in Figure \ref{fig:SOOD_imagenet_fw}. As a seed, we gave a set of 106 super-classes $\mathbb{C}$, selected from other datasets~\cite{kuznetsova2020openimages, pascal}. 
\subsection{Image Classification Data Engine}
\label{subsec:image_class_engine}
To create SOOD-ImageNet-C, we began from ImageNet21K-P~\cite{imagenet21k-p}, a dataset comprising around 11K classes or synsets which can be easily organized in a semantic hierarchical structure. For this purposes, we relied on WordNet~\cite{wordnet} database, which contains more than 100K words and their semantic relationships. Specifically, WordNet defines hyponyms—words with more specific meanings than a general term (\eg "apple" is a hyponym of "fruit"). For each category in ImageNet21K-P, we extracted its list of hyponyms. By navigating this structure, we could retrieve hyponyms of each category at different semantic levels. For example, exploring the hyponyms of "apple" allows us to associate "Golden Delicious" with "fruit". Using this hierarchical structure, we gathered all synsets in ImageNet21K-P that are semantically related to a specific super-class $\mathcal{C} \in \mathbb{C}$ of SOOD-ImageNet. We denote all synsets associated with $\mathcal{C}$ as $\mathcal{S}_{\mathcal{C}}$, representing sub-classes of $\mathcal{C}$. When a super-class $\mathcal{C}$ was missing from the hierarchy, we used Sentence Transformer~\cite{reimers-2019-sentence-transformer} to find synsets with the closest meaning and their hyponyms. A synset was considered close to $\mathcal{C}$ if the similarity score given by Sentence Transformer was higher than a threshold $t \in [0, 1]$. We empirically set the threshold to $0.5$ to keep a good balance between semantic correlation and the size of $\mathcal{S}_{\mathcal{C}}$. However, considering natural language semantic relationships alone is insufficient for computer vision tasks. For instance, a hyponym of "fruit" is "seed," which is not visually correlated with the concept of fruit. Given the dataset's size after the first processing stage—over 6M images and more than 6K synsets—we performed two filtering steps, A and B. Step A aimed to remove as many spurious synsets as possible. Leveraging the rise of large Vision-Language Models (VLMs)~\cite{vlm_survey}, we used a pre-trained model for Visual Question Answering, specifically PaliGemma~\cite{beyer2024paligemma}. For each synset $s_{\mathcal{C}} \in \mathcal{S}_{\mathcal{C}}$, we prompted the VLM with a grid of 16 images, randomly sampled from $s$, and a textual query: «{\textit{Do all the 16 images contain $\mathcal{C}$? y/n}}». If the answer was yes, we retained the synset $s_{\mathcal{C}}$; otherwise, we discarded it. After step A, we obtained a dataset comprising more than 2,000 synsets, accounting for around 2 million images. In step B, we performed a human check of the remaining synsets. The settings were similar to step A: 16 examples for each synset $s_{\mathcal{C}} \in \mathcal{S}_{\mathcal{C}}$ were shown to a human annotator, who discarded synsets not visually correlated with the class $\mathcal{C}$. We finally removed all duplicates, ensuring $\mathcal{S}_{\mathcal{C}i} \cap \mathcal{S}_{\mathcal{C}j} = \emptyset \quad \forall ; i \neq j$. We also decided to keep only classes with $|\mathcal{S}_{\mathcal{C}}| \ge 10$ to ensure sufficient samples and variability.

At this stage, we had reorganized ImageNet21K-P~\cite{imagenet21k-p} into a set of 56 super-classes $\mathbb{C}$, each with at least 10 corresponding sub-classes $\mathcal{S}_{\mathcal{C}} = \{ s_
{\mathcal{C}} \mid s_{\mathcal{C}} \subset \mathcal{C} \}$. For the scoring step, we need to defined $p$, the measure of the semantic correlation between the class and each sub-class to determine what is out-of-distribution. As discussed earlier, we needed to account for the visual content of the images for computer vision tasks. Therefore, we employed CLIP~\cite{clip} embeddings to measure the correlation between the class $\mathcal{C}$ and its sub-classes $\mathcal{S}_{\mathcal{C}}$. The correlation score of each sub-class, $p(s_{\mathcal{C}})$, is the average similarity of each image $im \in s_{\mathcal{C}}$ with the class $\mathcal{C}$, computed as the normalized cosine similarity of the corresponding CLIP embeddings. As aforementioned, the scoring function $p$ theoretically allows splitting $\mathbb{C}$ into an arbitrary number of partitions. However, considering real data, the partitions must contain enough data to train and test models. Moreover, testing large models on many different subsets consumes time and resources. Therefore, even though studying the influence of coarse and fine-grained partitions would be interesting, we limited the number of partitions to three for this exploratory work. Specifically, we considered in-distribution all sub-classes within the 60th percentile, i.e., $\mathcal{C}_{iid} = \{ s_{\mathcal{C}} \mid p(s_{\mathcal{C}}) > 0.4 \}$, using them as the training set. The rest was further split into two test sets: an easy test $\mathcal{C}_{ood}^{E} = \{ s_{\mathcal{C}} \mid 0.2 < p(s_{\mathcal{C}}) < 0.4 \}$ and a hard test $\mathcal{C}_{ood}^{H} = \{ s_{\mathcal{C}} \mid p(s_{\mathcal{C}}) < 0.2 \}$. By dividing the OOD test set into two partitions, we can evaluate performance with gradually more challenging examples. 
\subsection{Semantic Segmentation Data Engine}
\label{sec:sem_seg_engine}
Considering the dataset SOOD-ImageNet-C created through the procedure described in Section \ref{subsec:image_class_engine}, our focus shifted to obtaining the SOOD-ImageNet-S dataset for semantic segmentation using the splits obtained from the Image Classification Data Engine: $\mathcal{C}_{iid}$, $\mathcal{C}^{E}_{ood}$, and $\mathcal{C}^{H}_{ood}$. Building a large and high-quality annotated dataset for semantic segmentation is both costly and time-consuming. However, as it was largely proved, using weak labels of large dataset can significantly improve segmentation models performance, surpassing fully-supervised models with smaller dataset~\cite{yi2021learning}. Therefore, to annotate our training dataset $\mathcal{C}_{iid}$, we employed the CLIPSeg pre-trained model~\cite{luddecke2022image}, which generates image segmentations based on its super-class as the textual prompt. This approach allowed us to obtain efficiently weak labels of approximately 1 million images balancing quality and quantity. On the other hand, to ensure the most precise annotations possible for both the easy test $\mathcal{C}^{E}_{ood}$ and the hard test $\mathcal{C}^{H}_{ood}$, we randomly selected 120 images from each of the 56 classes within $\mathcal{C}^{E}_{ood}$ and $\mathcal{C}^{H}_{ood}$ for annotation. The 13440 chosen images where processed using a more robust and precise foundation model to achieve scalable annotations. A human verification step was then performed to maintain quality and avoid labelling errors. For the initial annotation, we utilized a combination of Grounding Dino~\cite{liu2023grounding} and SAM 2~\cite{ravi2024sam}. Grounding Dino was employed to identify the bounding box within each image based on the textual prompt corresponding to its class. The resulting bounding box was then provided as the input prompt to SAM 2, the latest foundation model for promptable visual segmentation.

\begin{figure}[tb]
  \centering
    \includegraphics[width=\linewidth]{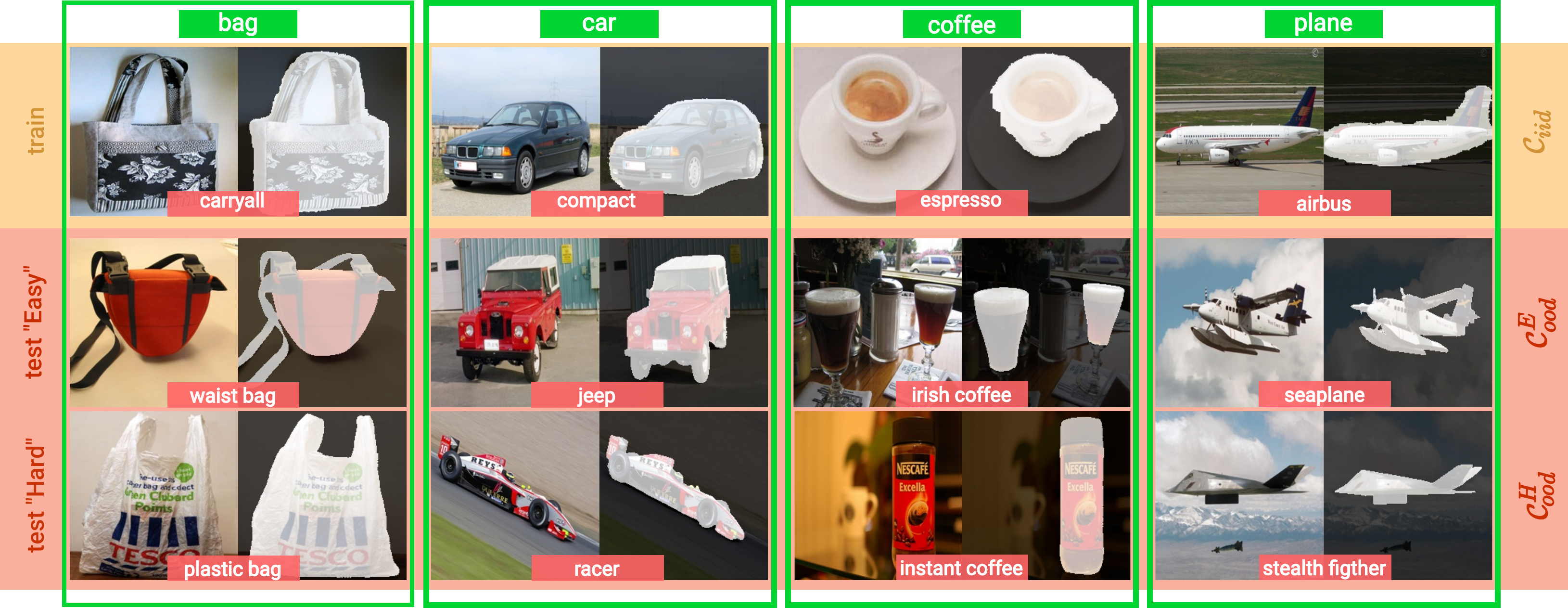}
    \caption{Example images from SOOD-ImageNet. Four classes (\ie bag, car, coffee and plane) with the corresponding classification (green) and segmentation labels are represented. It also possible to appreciate the semantic shift between $\mathcal{C}_{iid}$ (train), $\mathcal{C}^{E}_{ood}$ (test "Easy") and $\mathcal{C}^{H}_{ood}$ (test "Hard"). The original Imagenet-21K-P classes of the samples are reported in red.}
  \label{fig:dataset_label_examples}
\end{figure}

This process resulted in 13440 annotated images for the $\mathcal{C}^{E}_{ood}$ and $\mathcal{C}^{H}_{ood}$ datasets. Each image and its corresponding mask were subsequently reviewed by a human annotator, who discarded incorrectly annotated images. This process was crucial to remove incorrect annotations and ensure high-quality labeling for the test set.
After this verification step, the final test set comprised a total of 8335 images, with 4385 images in the easy test ($\mathcal{C}^{E}_{ood}$) and 3950 images in the hard test ($\mathcal{C}^{H}_{ood}$). Some samples of the images and the corresponding labels are shown in Figure~\ref{fig:dataset_label_examples}. It is possible to appreciate the semantic shift both in the language and vision domains.
\section{Experiments and Discussion}
\label{sec:experiment}
To motivate the need for introducing an SOOD generalization benchmark, we conducted a series of experiments using both SOOD-ImageNet-C and SOOD-ImageNet-S. By testing various models and augmentation techniques, our aim is to demonstrate that SOOD generalization remains an open challenge in computer vision and current technologies are insufficient to address this issue, highlighting the necessity for the development of specialized, ad-hoc solutions.
\begin{table}[tb]
  \caption{Results from the tests performed on the "Easy" $\mathcal{C}^{E}_{ood}$ and the "Hard" $\mathcal{C}^{H}_{ood}$ sets for image classification. The loss of performance between "Easy" and "Hard" is highlighted in the third column. For each model, the size in millions of parameters and the architecture of the backbone are reported.}
\begin{minipage}{\textwidth}
    \centering
    \setlength{\tabcolsep}{2pt}
    \begin{tabular}{@{}lcccccc@{}}
      \toprule
       Model & \hspace{0.75pt} $\mathcal{C}^{E}_{ood}$~(F1) & \hspace{0.75pt} $\mathcal{C}^{H}_{ood}$~(F1) & \hspace{0.75pt} $\Delta$ (\%) & \hspace{0.75pt} Size (M) & \hspace{0.75pt} CNN & \hspace{0.75pt} Transformer \\
      \midrule
      EfficientNet~\cite{koonce2021efficientnet} (B6) & 0.65 & 0.49 & -24.38 & 40.9 & $\times$ &  \\ 
      MaxxViT-V2~\cite{tu2022maxvit} (L) & 0.43 & 0.32 & -25.75 & 214.2 & $\times$ & $\times$ \\ 
      MobileNetV3~\cite{Howard2019SearchingFM} & 0.63 & 0.47 & -25.99 & 4.3 & $\times$ &  \\ 
      ResNet50~\cite{resnet} & 0.62 & 0.46 & -26.80 & 23.6 & $\times$ &  \\ 
      Swinv2~\cite{liu2022swin} (B) & 0.62 & 0.46 & -25.56 & 86.9 &  & $\times$ \\ 
      Swinv2~\cite{liu2022swin} (L) & 0.62 & 0.46 & -26.04 & 195.2 &  & $\times$ \\ 
      ViT~\cite{dosovitskiy2020image} (B) & 0.54 & 0.39 & -27.82 & 85.8 &  & $\times$ \\ 
      ViT~\cite{dosovitskiy2020image} (L) & 0.25 & 0.19 & -24.12 & 303.4 &  & $\times$ \\ 
      \hdashline
      BLIP~\cite{li2022blip} & 0.07 & 0.06 & -16.21 & 384.6 &  & $\times$ \\ 
      Florence2~\cite{xiao2024florence} & 0.19 & 0.16 & -18.13 & 822.7 &  & $\times$ \\ 
      InterNVL2~\cite{chen2024internvl} & 0.44 & 0.35 & -21.01 & 2205.7 &  & $\times$ \\ 
      LLaVa~\cite{liu2024visual} & 0.46 & 0.35 & -23.16 & 3663.9 &  & $\times$ \\ 
      \bottomrule
    \end{tabular}
    \label{tab:easy_hard_cls_test}
  \end{minipage}
\end{table}
\subsubsection{Experiments on Classification.} 
\begin{figure}[tb]
  \centering
  \begin{subfigure}{0.48\linewidth}
    \includegraphics[width=\linewidth]{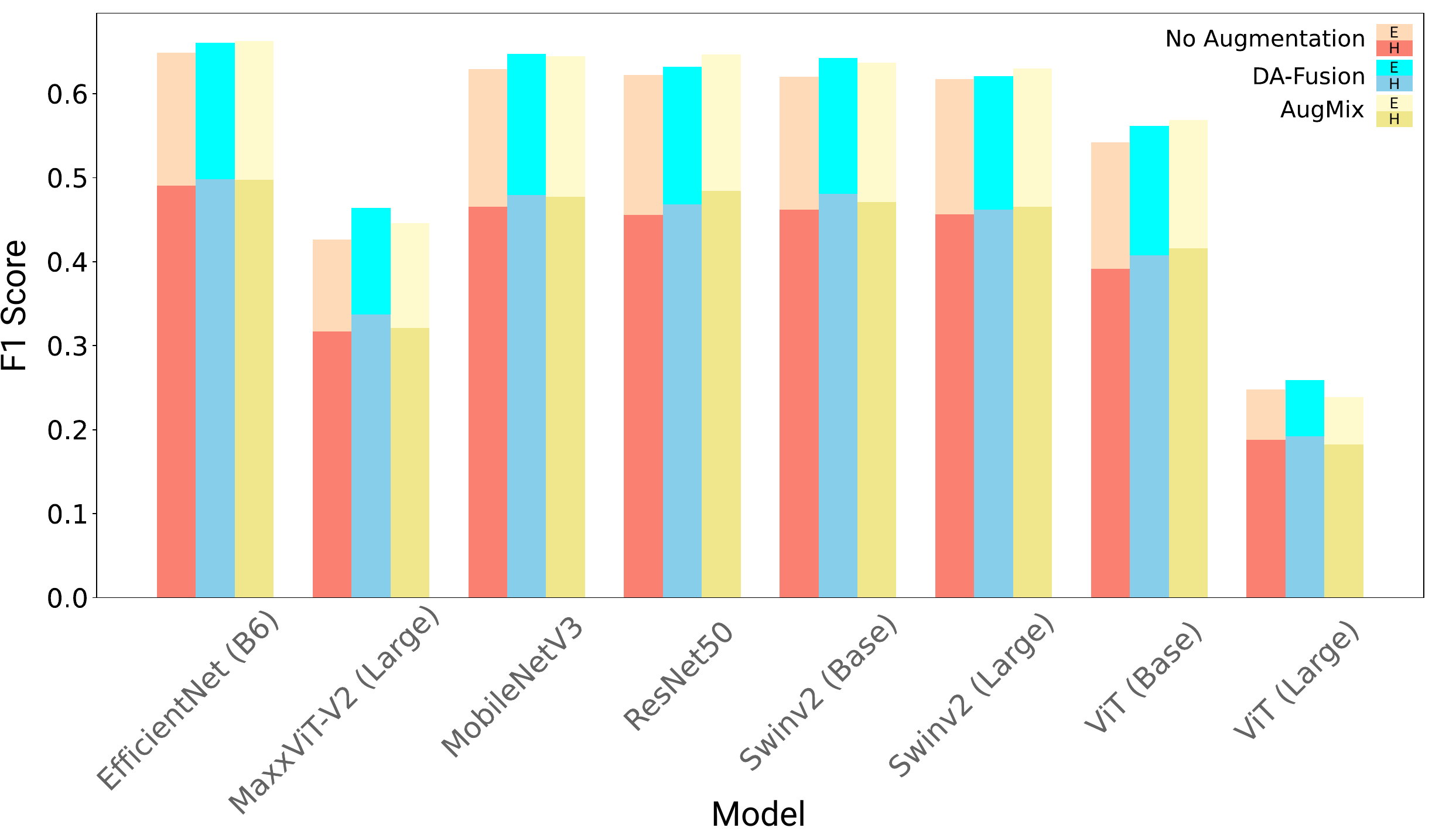}
    \caption{Classification}
    \label{fig:augmentations_cls}
  \end{subfigure}
  \hfill
  \begin{subfigure}{0.48\linewidth}
    \includegraphics[width=\linewidth]{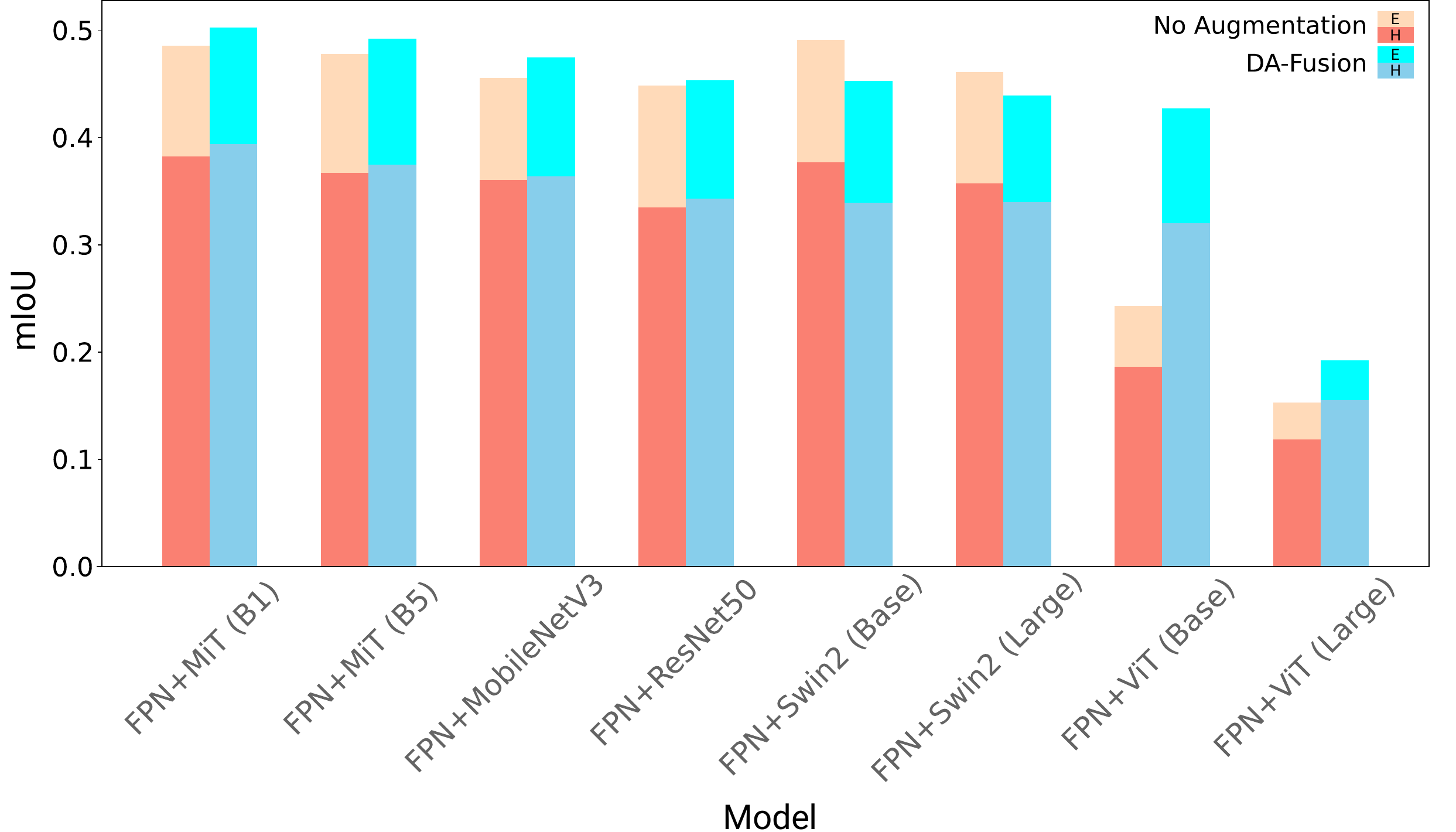}
    \caption{Segmentation}
    \label{fig:augmentations_seg}
  \end{subfigure}
  \caption{The histograms represent the impact of different data augmentation methods, \ie AugMix~\cite{2020augmix} and DA-Fusion~\cite{trabucco2024-dafusion}, on SOOD generalization of models. Performance on $\mathcal{C}^{E}_{ood}$ (Easy) are represented with lighter shades. }
  \label{fig:augmentations}
\end{figure}

To conduct experiments on SOOD-ImageNet-C, we selected eight popular backbones for classification, including both convolutional and transformer architectures, each paired with a dense classifier. All models were trained from scratch on images from $\mathcal{C}_{iid}$ without pre-training, ensuring the validity of the OOD hypothesis. The models were trained for 25 epochs using a base learning rate of $1.25\mathrm{e}{-3}$ with cosine annealing scheduling~\cite{loshchilov2017sgdr_cosine_lr} and a weight decay of $5\mathrm{e}{-3}$. We employed cross-entropy loss and the AdamW optimizer, with a batch size of 48. After training, the models were tested separately on $\mathcal{C}^{E}_{ood}$ and $\mathcal{C}^{H}_{ood}$. We evaluated the performance of each model using the average F1 score, as shown in Table~\ref{tab:easy_hard_cls_test}. The third column reveals that the models perform worse on $\mathcal{C}^{H}_{ood}$ compared to $\mathcal{C}^{E}_{ood}$, which is semantically closer to the training set, with at least a 24\% drop in F1 score. This indicates that semantic shifts significantly affect model performance, and testing on IID data alone is insufficient to assess whether a model generalizes well to certain classes. Interestingly, the size and architecture of the models do not seem to have a significant impact on the results, with performance degradation being relatively consistent across different architectures, as similarly reported by Zhao et al.\cite{ood-cv}. However, a closer examination shows that convolutional architectures\cite{koonce2021efficientnet, tu2022maxvit, Howard2019SearchingFM, resnet} and SwinV2~\cite{liu2022swin} tend to perform better in absolute terms. These designs leverage visual priors such as hierarchy, translation invariance, and locality, unlike ViT architectures. We hypothesize that these priors contribute to better SOOD generalization. On the other hand, ViT architectures—especially the "Large" models—may require more data and longer training times to achieve convergence. Further experiments are needed to validate this hypothesis.
\begin{figure}[tb]
  \centering
    \includegraphics[width=\linewidth]{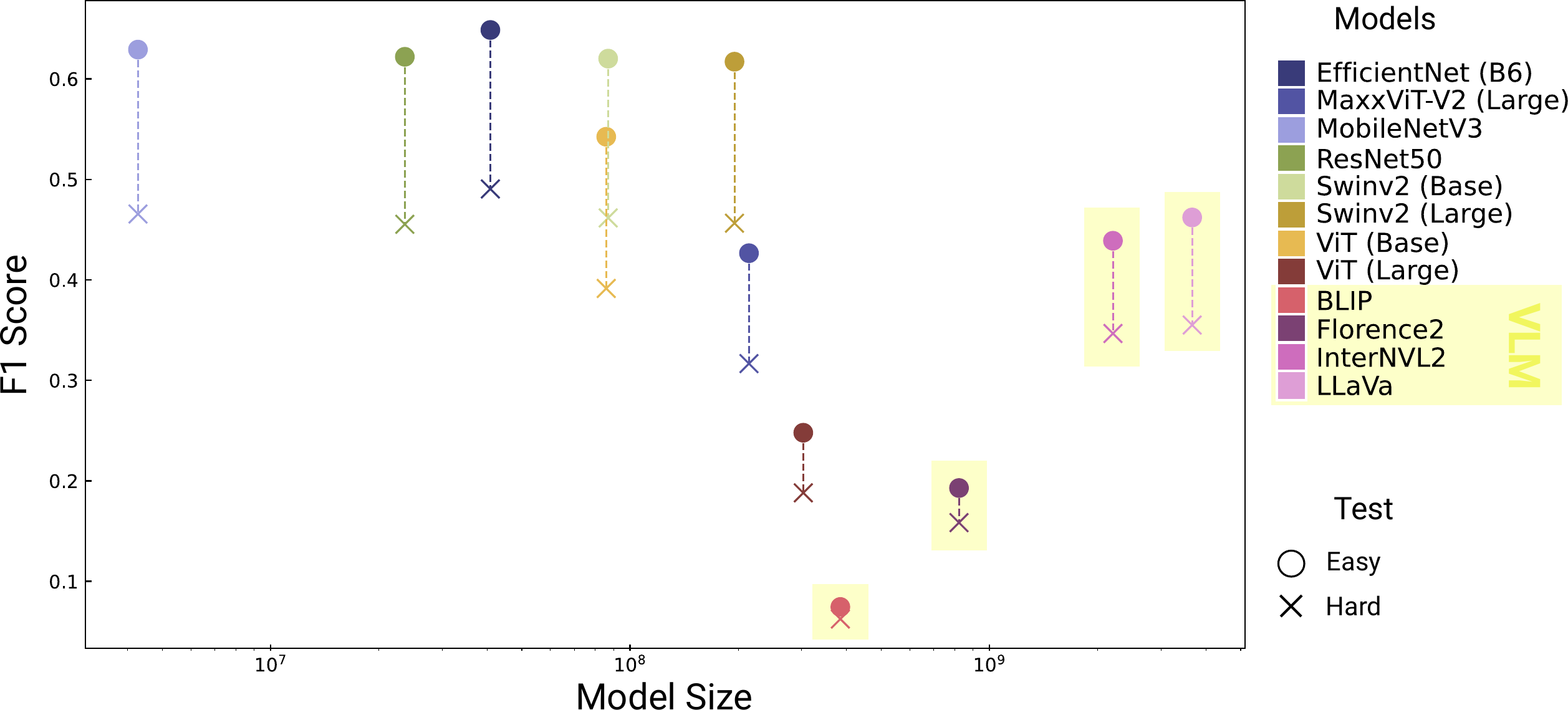}
    \caption{The graph compares various models in terms of performance (F1 score) and number of parameters, when tested on $\mathcal{C}^{E}_{ood}$ (Easy) and $\mathcal{C}^{H}_{ood}$ (Hard) for image classification. We highlighted the gap between Easy and Hard tests with a dotted line. Pre-trained VLMs are in yellow.}
  \label{fig:comparison_cls}
\end{figure}
A relevant question at this point is whether standard techniques are effective in improving SOOD generalization. A very popular approach to improve generalization of DL models is data augmentation. Thus, we applied AugMix~\cite{2020augmix}, a recent technique that combines random transformations with a consistency loss. We selected AugMix because the authors have reported its success in handling distributional shifts during testing. Additionally, we implemented the recently released DA-Fusion~\cite{trabucco2024-dafusion}, which leverages Stable Diffusion~\cite{Rombach_stable_diffusion} to generate semantically augmented samples. We hypothesized that DA-Fusion would be particularly well-suited to address the SOOD generalization problem due to its focus on semantic augmentation. However, as shown in Figure~\ref{fig:augmentations_cls}, neither technique produced a significant improvement when tested on SOOD data. This suggests that even state-of-the-art data augmentation techniques are insufficient to address the SOOD generalization problem.

For the sake of completeness, we also tested some state-of-the-art foundation models on $\mathcal{C}^{E}_{ood}$ and $\mathcal{C}^{H}_{ood}$, including BLIP~\cite{li2022blip}, Florence2~\cite{xiao2024florence}, InternVL2~\cite{chen2024internvl}, and LLaVa~\cite{liu2024visual}. Before delving into the analysis, it is important to acknowledge a key consideration: defining what constitutes OOD for a VLM (Vision-Language Model) pre-trained on large-scale datasets is challenging because it is difficult to guarantee that the model has never encountered a particular condition during training. Nevertheless, studying the behavior of these models in an SOOD context is still informative. To use a VLM as a classifier, we prompted the model with the testing image and the textual prompt: «{\textit{Describe what is in the image in one word.}}» We considered the answer correct if it matched the ground truth super-class or sub-class name. Results are presented in Table~\ref{tab:easy_hard_cls_test} and visually depicted in Figure~\ref{fig:comparison_cls}. Interestingly, even highly capable VLMs underperformed compared to standard vision models without pre-training. This discrepancy may partly stem from the strict evaluation criteria, as VLMs occasionally generated synonyms of the ground truth class name. Despite this, the performance gap between $\mathcal{C}^{E}{ood}$ and $\mathcal{C}^{H}{ood}$ remains significant, ranging from 16\% to 23\%, underscoring that even VLMs are susceptible to semantic shifts.
\subsubsection{Experiments on Semantic Segmentation.} 
To validate SOOD-ImageNet-S, we followed a pipeline similar to the one used for SOOD-ImageNet-C. We selected eight models for semantic segmentation~\cite{torchseg}, combining a Feature Pyramid Network (FPN)~\cite{fpn} with various relevant backbones. As before, we included backbones of different sizes and architectures. The training settings were identical to those used for SOOD-ImageNet-C. We evaluated performance using the mean Intersection over Union (mIoU), the most widely used metric for semantic segmentation.
\begin{table}[tb]
  \caption{Results from the tests performed on the "Easy" $\mathcal{C}^{E}_{ood}$ and the "Hard" $\mathcal{C}^{H}_{ood}$ sets for semantic segmentation. The loss of performance between "Easy" and "Hard" is highlighted in the third column. For each model, the size in millions of parameters and the architecture of the backbone are reported.}
  \begin{minipage}{\textwidth}
    \centering
    \begin{tabular}{@{}lcccccc@{}}
      \toprule
      Model Name & \hspace{0.5pt} $\mathcal{C}^{E}_{ood}$~(mIoU) & \hspace{0.5pt} $\mathcal{C}^{H}_{ood}$~(mIoU) & \hspace{0.5pt} $\Delta$ (\%) & \hspace{0.5pt} Size (M) & \hspace{0.75pt} CNN & \hspace{0.5pt} Transformer \\
      \midrule
      MiT~\cite{xie2021segformer} (B1) & 0.49 & 0.38 & -21.25 & 20.2 &  & $\times$  \\ 
      MiT~\cite{xie2021segformer} (B5) & 0.48 & 0.37 & -23.17 & 88.5 &  & $\times$  \\ 
      MobileNetV3~\cite{Howard2019SearchingFM} & 0.46 & 0.36 & -20.92 & 10.1 & $\times$ &  \\ 
      ResNet50~\cite{resnet} & 0.45 & 0.33 & -25.35 & 32.0 & $\times$ &  \\ 
      Swin2~\cite{liu2022swin} (B) & 0.49 & 0.38 & -23.27 & 94.4 &  & $\times$ \\ 
      Swin2~\cite{liu2022swin} (L) & 0.46 & 0.36 & -22.55 & 203.1 &  & $\times$ \\ 
      ViT~\cite{dosovitskiy2020image} (B) & 0.24 & 0.19 & -23.31 & 94.7 &  & $\times$ \\ 
      ViT~\cite{dosovitskiy2020image} (L) & 0.15 & 0.12 & -22.77 & 313.0 &  & $\times$ \\
      \hdashline
      GroupViT~\cite{xu2022groupvit} & 0.43 & 0.39 & -9.19 & 55.1 &  & $\times$ \\ 
      Florence2~\cite{xiao2024florence} & 0.54 & 0.50 & -8.41 & 822.7 &  & $\times$ \\ 
      FastSAM~\cite{zhao2023fast} & 0.31 & 0.28 & -8.88 & 68.0 &  & $\times$ \\ 
      \bottomrule
    \end{tabular}
    \label{tab:easy_hard_seg_test}
  \end{minipage}
\end{table}
As expected, the results aligned with our observations for SOOD-ImageNet-C. In Table~\ref{tab:easy_hard_seg_test}, we see a performance drop between 21\% and 25\%. The different models exhibited similar behavior, confirming that SOOD generalization is an open challenge also in the semantic segmentation task. We also applied data augmentation techniques in the context of semantic segmentation. Since AugMix is applied at training time, it was not possible ensure the alignment of the segmentation masks with the augmented image in an efficient way. In contrast, DA-Fusion allows for the generation of augmented samples offline. This approach enabled us to extract segmentation masks using CLIPSeg, as described in Section~\ref{sec:sem_seg_engine}. For this reason, we only used DA-Fusion to augment SOOD-ImageNet-S. The results, depicted in Figure~\ref{fig:augmentations_seg} reflect the observations made earlier, with the exception of ViT, which appears to benefit significantly from data augmentation in the segmentation task. However, the gap between $\mathcal{C}^{E}_{ood}$ and $\mathcal{C}^{H}_{ood}$ remains.

\begin{figure}[tb]
  \centering
    \includegraphics[width=\linewidth]{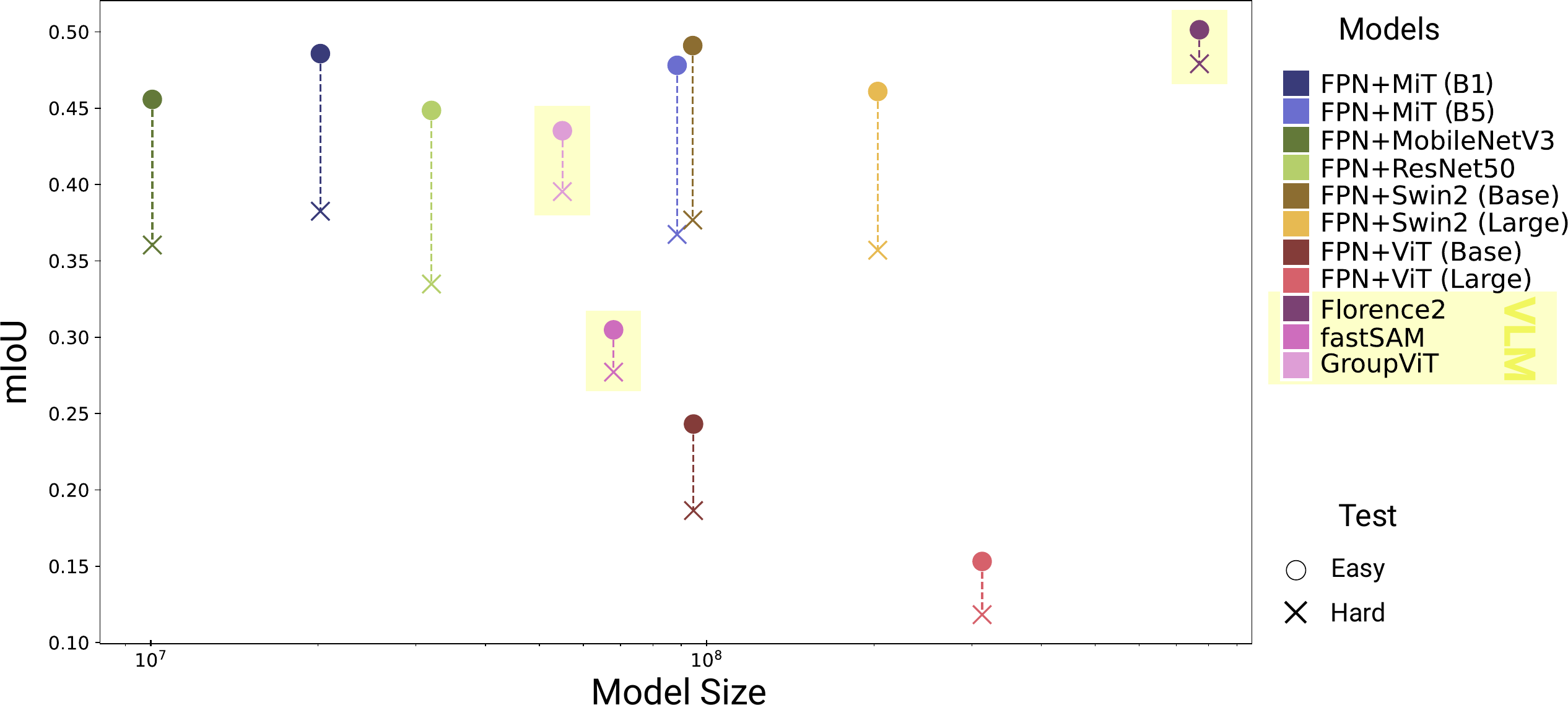}
    \caption{The graph compares various models in terms of performance (mIoU) and number of parameters, when tested on $\mathcal{C}^{E}_{ood}$ (Easy) and $\mathcal{C}^{H}_{ood}$ (Hard) for semantic segmentation. We highlighted the gap between Easy and Hard tests with a dotted line. Pre-trained VLMs are in yellow.}
  \label{fig:comparison_seg}
\end{figure}

For the segmentation task, we selected a different set of foundation models. We retained Florence2~\cite{xiao2024florence} due to its multitask nature, and added FastSAM~\cite{zhao2023fast} and GroupViT~\cite{xu2022groupvit}, which are VLMs optimized for semantic segmentation. The results are presented in Figure~\ref{fig:comparison_seg}. In these experiments, we observed a notable difference compared to the classification task. Specifically, the gap between $\mathcal{C}^{E}_{ood}$ and $\mathcal{C}^{H}_{ood}$ was smaller. We speculate that this is because these VLMs natively support semantic segmentation, whereas in the classification experiments, we had to implement a workaround. This suggests that VLMs may have a better ability to address the SOOD generalization challenge, particularly Florence2, which demonstrated the best SOOD generalization capability even in the classification experiments. However, as previously discussed, it remains challenging to conduct a proper OOD test for a foundation model.
\section{Conclusion and Future Works}
\label{sec:conclusion}
In this work, we explore the problem of SOOD generalization, which refers to a model's ability to generalize to semantic shifts from the training classes. To the best of our knowledge, this problem is not well-studied in the literature. To address this gap, we introduce a novel dataset called SOOD-ImageNet. The dataset includes approximately 1.6 million images across 56 classes and is labeled for both image classification and semantic segmentation, making it one of the largest datasets in the OOD research domain. The dataset was generated using a custom data engine that innovatively integrates visual and language cues within a unique pipeline. It strikes a balance between quality, scalability, and flexibility—allowing third parties to tune the engine and customize their own datasets. Our experiments demonstrate that state-of-the-art DL models -- even large VLMs -- struggle with semantic shifts, and that SOOD generalization cannot be effectively addressed using standard techniques like data augmentation. We believe this work could foster further research on OOD in computer vision, particularly concerning SOOD generalization. As a pioneering effort in SOOD generalization, this work opens many avenues for future research. It is important to note that SOOD-ImageNet currently covers less than 15\% of the original ImageNet-21K-P. There are two main reasons for this. First, the choice of seed classes may not fully capture the diversity of ImageNet-21K-P. A more deliberate selection of seed classes could improve coverage. Second, we observed that in some cases, the Sentence Transformer fails to associate relevant sub-classes. For example, certain animal species may be excluded because they are denoted by scientific names that the language model does not recognize. Enhancing this component of our data engine could also improve coverage. Expanding the size of the dataset will enable more extensive experiments, such as conducting finer-grained evaluations and using more than two partitions for testing. Moreover, models with hundreds of millions of parameters would benefit from a larger number of examples. It would be also interesting to explore the potential of using SOOD-ImageNet as a benchmark for open set recognition (OSR). Specifically, $\mathcal{C}^{E}{ood}$ (Easy) and $\mathcal{C}^{H}{ood}$ (Hard) could serve as "unknown" classes, providing a valuable opportunity to further test and refine OSR techniques.
\section*{Acknowledgements}
This work was supported by the Italian Minister for University and Research (MUR) under the initiative “PON Ricerca e Innovazione 2014 - 2020”, CUP C95F21007870007 and by the European Union’s Horizon 2020 research and innovation programme under the grant agreement No. 101006732 (DrapeBot).

\bibliographystyle{splncs04}
\bibliography{main}

\begin{thebibliography}{10}
\providecommand{\url}[1]{\texttt{#1}}
\providecommand{\urlprefix}{URL }
\providecommand{\doi}[1]{https://doi.org/#1}

\bibitem{Angarano_2024_CVPR_agriculture}
Angarano, S., Martini, M., Navone, A., Chiaberge, M.: Domain generalization for crop segmentation with standardized ensemble knowledge distillation. In: IEEE Conf. Comput. Vis. Pattern Recog. Worksh. pp. 5450--5459 (2024)

\bibitem{pmlr-v220-bashkirova23a_visda}
Bashkirova, D., Mishra, S., Lteif, D., Teterwak, P., Kim, D., Alladkani, F., Akl, J., Calli, B., Bargal, S.A., Saenko, K., Kim, D., Seo, M., Jeon, Y., Choi, D.G., Ettedgui, S., Giryes, R., Abu-Hussein, S., Xie, B., Li, S.: Visda 2022 challenge: Domain adaptation for industrial waste sorting. In: Proceedings of the NeurIPS 2022 Competitions Track. pp. 104--118 (2022)

\bibitem{beyer2024paligemma}
Beyer, L., Steiner, A., Pinto, A.S., Kolesnikov, A., Wang, X., Salz, D., Neumann, M., Alabdulmohsin, I., Tschannen, M., Bugliarello, E., et~al.: Paligemma: A versatile 3b vlm for transfer. arXiv preprint arXiv:2407.07726  (2024)

\bibitem{chen2024internvl}
Chen, Z., Wu, J., Wang, W., Su, W., Chen, G., Xing, S., Zhong, M., Zhang, Q., Zhu, X., Lu, L., et~al.: Internvl: Scaling up vision foundation models and aligning for generic visual-linguistic tasks. In: IEEE Conf. Comput. Vis. Pattern Recog. pp. 24185--24198 (2024)

\bibitem{texture}
Cimpoi, M., Maji, S., Kokkinos, I., Mohamed, S., Vedaldi, A.: Describing textures in the wild. In: IEEE Conf. Comput. Vis. Pattern Recog. pp. 3606--3613 (2014)

\bibitem{Conde_2021_CVPR_clip_art}
Conde, M.V., Turgutlu, K.: Clip-art: Contrastive pre-training for fine-grained art classification. In: IEEE Conf. Comput. Vis. Pattern Recog. Worksh. pp. 3956--3960 (2021)

\bibitem{torchseg}
Corley, I.: Torchseg. \url{https://github.com/isaaccorley/torchseg}

\bibitem{imagenet_cvpr09}
Deng, J., Dong, W., Socher, R., Li, L.J., Li, K., Fei-Fei, L.: {ImageNet: A Large-Scale Hierarchical Image Database}. In: IEEE Conf. Comput. Vis. Pattern Recog. (2009)

\bibitem{dosovitskiy2020image}
Dosovitskiy, A., Beyer, L., Kolesnikov, A., Weissenborn, D., Zhai, X., Unterthiner, T., Dehghani, M., Minderer, M., Heigold, G., Gelly, S., Uszkoreit, J., Houlsby, N.: An image is worth 16x16 words: Transformers for image recognition at scale. In: Int. Conf. Learn. Represent. (2021)

\bibitem{pascal}
Everingham, M., Eslami, S.M.A., Van~Gool, L., Williams, C.K.I., Winn, J., Zisserman, A.: The pascal visual object classes challenge: A retrospective. Int. J. Comput. Vis.  \textbf{111},  98--136 (2015)

\bibitem{filos2020can}
Filos, A., Tigkas, P., McAllister, R., Rhinehart, N., Levine, S., Gal, Y.: Can autonomous vehicles identify, recover from, and adapt to distribution shifts? In: Int. Conf. Mach. Learn. pp. 3145--3153. PMLR (2020)

\bibitem{resnet}
He, K., Zhang, X., Ren, S., Sun, J.: Deep residual learning for image recognition. In: IEEE Conf. Comput. Vis. Pattern Recog. pp. 770--778 (2016)

\bibitem{imagenet-r}
Hendrycks, D., Basart, S., Mu, N., Kadavath, S., Wang, F., Dorundo, E., Desai, R., Zhu, T., Parajuli, S., Guo, M., Song, D., Steinhardt, J., Gilmer, J.: The many faces of robustness: A critical analysis of out-of-distribution generalization. In: Int. Conf. Comput. Vis. pp. 8340--8349 (2021)

\bibitem{imagenet-c}
Hendrycks, D., Dietterich, T.: Benchmarking neural network robustness to common corruptions and perturbations. In: Int. Conf. Learn. Represent. (2019)

\bibitem{2020augmix}
Hendrycks*, D., Mu*, N., Cubuk, E.D., Zoph, B., Gilmer, J., Lakshminarayanan, B.: Augmix: A simple method to improve robustness and uncertainty under data shift. In: Int. Conf. Learn. Represent. (2020), \url{https://openreview.net/forum?id=S1gmrxHFvB}

\bibitem{imagenet-a}
Hendrycks, D., Zhao, K., Basart, S., Steinhardt, J., Song, D.: Natural adversarial examples. In: IEEE Conf. Comput. Vis. Pattern Recog. pp. 15262--15271 (2021)

\bibitem{inat}
Horn, G.V., Cole, E., Beery, S., Wilber, K., Belongie, S., MacAodha, O.: Benchmarking representation learning for natural world image collections. In: IEEE Conf. Comput. Vis. Pattern Recog. pp. 12879--12888 (2021)

\bibitem{Howard2019SearchingFM}
Howard, A.G., Sandler, M., Chu, G., Chen, L.C., Chen, B., Tan, M., Wang, W., Zhu, Y., Pang, R., Vasudevan, V., Le, Q.V., Adam, H.: Searching for mobilenetv3. Int. Conf. Comput. Vis. pp. 1314--1324 (2019)

\bibitem{mos}
Huang, R., Li, Y.: Mos: Towards scaling out-of-distribution detection for large semantic space. In: IEEE Conf. Comput. Vis. Pattern Recog. pp. 8710--8719 (2021)

\bibitem{Khandelwal_2022_CVPR_clip_embodAI}
Khandelwal, A., Weihs, L., Mottaghi, R., Kembhavi, A.: Simple but effective: Clip embeddings for embodied ai. In: IEEE Conf. Comput. Vis. Pattern Recog. pp. 14829--14838 (2022)

\bibitem{krizhevsky2009learning}
Krizhevsky, A.: Learning multiple layers of features from tiny images. Master's thesis, University of Toronto  (2009)

\bibitem{kuznetsova2020openimages}
Kuznetsova, A., Rom, H., Alldrin, N., Uijlings, J., Krasin, I., Pont-Tuset, J., Kamali, S., Popov, S., Malloci, M., Kolesnikov, A., et~al.: The open images dataset v4: Unified image classification, object detection, and visual relationship detection at scale. Int. J. Comput. Vis.  \textbf{128},  1956--1981 (2020)

\bibitem{iNatural2021-osr}
Lang, N., Sn{\ae}bjarnarson, V., Cole, E., Mac~Aodha, O., Igel, C., Belongie, S.: From coarse to fine-grained open-set recognition. In: IEEE Conf. Comput. Vis. Pattern Recog. pp. 17804--17814 (2024)

\bibitem{li2022blip}
Li, J., Li, D., Xiong, C., Hoi, S.: Blip: Bootstrapping language-image pre-training for unified vision-language understanding and generation. In: Int. Conf. Mach. Learn. pp. 12888--12900. PMLR (2022)

\bibitem{fpn}
Lin, T.Y., Dollar, P., Girshick, R., He, K., Hariharan, B., Belongie, S.: Feature pyramid networks for object detection. In: IEEE Conf. Comput. Vis. Pattern Recog. (July 2017)

\bibitem{liu2024visual}
Liu, H., Li, C., Wu, Q., Lee, Y.J.: Visual instruction tuning. Advances in neural information processing systems  \textbf{36} (2024)

\bibitem{liu2023grounding}
Liu, S., Zeng, Z., Ren, T., Li, F., Zhang, H., Yang, J., Li, C., Yang, J., Su, H., Zhu, J., et~al.: Grounding dino: Marrying dino with grounded pre-training for open-set object detection. arXiv preprint arXiv:2303.05499  (2023)

\bibitem{liu2022swin}
Liu, Z., Hu, H., Lin, Y., Yao, Z., Xie, Z., Wei, Y., Ning, J., Cao, Y., Zhang, Z., Dong, L., et~al.: Swin transformer v2: Scaling up capacity and resolution. In: IEEE Conf. Comput. Vis. Pattern Recog. pp. 12009--12019 (2022)

\bibitem{loshchilov2017sgdr_cosine_lr}
Loshchilov, I., Hutter, F.: {SGDR}: Stochastic gradient descent with warm restarts. In: Int. Conf. Learn. Represent. (2017)

\bibitem{luddecke2022image}
L{\"u}ddecke, T., Ecker, A.: Image segmentation using text and image prompts. In: IEEE Conf. Comput. Vis. Pattern Recog. pp. 7086--7096 (2022)

\bibitem{coco-o}
Mao, X., Chen, Y., Zhu, Y., Chen, D., Su, H., Zhang, R., Xue, H.: Coco-o: A benchmark for object detectors under natural distribution shifts. In: Int. Conf. Comput. Vis. pp. 6316--6327 (2023)

\bibitem{wordnet}
Miller, G.A.: Wordnet: a lexical database for english. Commun. ACM  \textbf{38}(11),  39–41 (1995)

\bibitem{clip}
Radford, A., Kim, J.W., Hallacy, C., Ramesh, A., Goh, G., Agarwal, S., Sastry, G., Askell, A., Mishkin, P., Clark, J., Krueger, G., Sutskever, I.: Learning transferable visual models from natural language supervision. In: Int. Conf. Mach. Learn. vol.~139, pp. 8748--8763 (2021)

\bibitem{ravi2024sam}
Ravi, N., Gabeur, V., Hu, Y.T., Hu, R., Ryali, C., Ma, T., Khedr, H., R{\"a}dle, R., Rolland, C., Gustafson, L., et~al.: Sam 2: Segment anything in images and videos. arXiv preprint arXiv:2408.00714  (2024)

\bibitem{reimers-2019-sentence-transformer}
Reimers, N., Gurevych, I.: Sentence-bert: Sentence embeddings using siamese bert-networks. In: Proceedings of the 2019 Conference on Empirical Methods in Natural Language Processing (2019)

\bibitem{imagenet21k-p}
Ridnik, T., Ben-Baruch, E., Noy, A., Zelnik, L.: Imagenet-21k pretraining for the masses. In: Proceedings of the Neural Information Processing Systems Track on Datasets and Benchmarks (2021)

\bibitem{Rombach_stable_diffusion}
Rombach, R., Blattmann, A., Lorenz, D., Esser, P., Ommer, B.: High-resolution image synthesis with latent diffusion models. In: IEEE Conf. Comput. Vis. Pattern Recog. pp. 10684--10695 (2022)

\bibitem{open-set-scheirer}
Scheirer, W.J., de~Rezende~Rocha, A., Sapkota, A., Boult, T.E.: Toward open set recognition. IEEE Trans. Pattern Anal. Mach. Intell.  \textbf{35}(7),  1757--1772 (2013)

\bibitem{koonce2021efficientnet}
Tan, M., Le, Q.: {E}fficient{N}et: Rethinking model scaling for convolutional neural networks. In: Int. Conf. Mach. Learn. vol.~97, pp. 6105--6114. PMLR (2019)

\bibitem{semantic_segmentation_survey}
Thisanke, H., Deshan, C., Chamith, K., Seneviratne, S., Vidanaarachchi, R., Herath, D.: Semantic segmentation using vision transformers: A survey. Engineering Applications of Artificial Intelligence  \textbf{126},  106669 (2023)

\bibitem{trabucco2024-dafusion}
Trabucco, B., Doherty, K., Gurinas, M.A., Salakhutdinov, R.: Effective data augmentation with diffusion models. In: Int. Conf. Learn. Represent. (2024), \url{https://openreview.net/forum?id=ZWzUA9zeAg}

\bibitem{tu2022maxvit}
Tu, Z., Talebi, H., Zhang, H., Yang, F., Milanfar, P., Bovik, A., Li, Y.: Maxvit: Multi-axis vision transformer. In: Eur. Conf. Comput. Vis. pp. 459--479. Springer (2022)

\bibitem{iNaturalist}
Van~Horn, G., Mac~Aodha, O., Song, Y., Cui, Y., Sun, C., Shepard, A., Adam, H., Perona, P., Belongie, S.: The inaturalist species classification and detection dataset. In: IEEE Conf. Comput. Vis. Pattern Recog. (2018)

\bibitem{vaze2022openset-semanticshift}
Vaze, S., Han, K., Vedaldi, A., Zisserman, A.: Open-set recognition: A good closed-set classifier is all you need. In: Int. Conf. Learn. Represent. (2022)

\bibitem{vim_openimage-o}
Wang, H., Li, Z., Feng, L., Zhang, W.: Vim: Out-of-distribution with virtual-logit matching. In: IEEE Conf. Comput. Vis. Pattern Recog. pp. 4921--4930 (2022)

\bibitem{pascal3d+}
Xiang, Y., Mottaghi, R., Savarese, S.: Beyond pascal: A benchmark for 3d object detection in the wild. In: IEEE Winter Conference on Applications of Computer Vision (WACV) (2014)

\bibitem{xiao2024florence}
Xiao, B., Wu, H., Xu, W., Dai, X., Hu, H., Lu, Y., Zeng, M., Liu, C., Yuan, L.: Florence-2: Advancing a unified representation for a variety of vision tasks. In: IEEE Conf. Comput. Vis. Pattern Recog. pp. 4818--4829 (2024)

\bibitem{xie2021segformer}
Xie, E., Wang, W., Yu, Z., Anandkumar, A., Alvarez, J.M., Luo, P.: Segformer: Simple and efficient design for semantic segmentation with transformers. Advances in neural information processing systems  \textbf{34},  12077--12090 (2021)

\bibitem{xu2022groupvit}
Xu, J., De~Mello, S., Liu, S., Byeon, W., Breuel, T., Kautz, J., Wang, X.: Groupvit: Semantic segmentation emerges from text supervision. In: IEEE Conf. Comput. Vis. Pattern Recog. pp. 18134--18144 (2022)

\bibitem{weed}
Xu, K., Shu, L., Xie, Q., Song, M., Zhu, Y., Cao, W., Ni, J.: Precision weed detection in wheat fields for agriculture 4.0: A survey of enabling technologies, methods, and research challenges. Computers and Electronics in Agriculture  \textbf{212},  108106 (2023)

\bibitem{Ye_2022_CVPR_ood_bench}
Ye, N., Li, K., Bai, H., Yu, R., Hong, L., Zhou, F., Li, Z., Zhu, J.: Ood-bench: Quantifying and understanding two dimensions of out-of-distribution generalization. In: IEEE Conf. Comput. Vis. Pattern Recog. pp. 7947--7958 (2022)

\bibitem{yi2021learning}
Yi, R., Huang, Y., Guan, Q., Pu, M., Zhang, R.: Learning from pixel-level label noise: A new perspective for semi-supervised semantic segmentation. IEEE Transactions on Image Processing  \textbf{31},  623--635 (2021)

\bibitem{zhang2024openood}
Zhang, J., Yang, J., Wang, P., Wang, H., Lin, Y., Zhang, H., Sun, Y., Du, X., Zhou, K., Zhang, W., Li, Y., Liu, Z., Chen, Y., Li, H.: Open{OOD} v1.5: Enhanced benchmark for out-of-distribution detection. In: NeurIPS 2023 Workshop on Distribution Shifts: New Frontiers with Foundation Models (2024)

\bibitem{vlm_survey}
Zhang, J., Huang, J., Jin, S., Lu, S.: Vision-language models for vision tasks: A survey. IEEE Trans. Pattern Anal. Mach. Intell.  \textbf{46}(8),  5625--5644 (2024)

\bibitem{nico++}
Zhang, X., He, Y., Xu, R., Yu, H., Shen, Z., Cui, P.: Nico++: Towards better benchmarking for domain generalization. In: IEEE Conf. Comput. Vis. Pattern Recog. pp. 16036--16047 (2023)

\bibitem{ood-cv}
Zhao, B., Yu, S., Ma, W., Yu, M., Mei, S., Wang, A., He, J., Yuille, A., Kortylewski, A.: Ood-cv: A benchmark for robustness to out-of-distribution shifts of individual nuisances in natural images. In: Eur. Conf. Comput. Vis. pp. 163--180 (2022)

\bibitem{zhao2023fast}
Zhao, X., Ding, W., An, Y., Du, Y., Yu, T., Li, M., Tang, M., Wang, J.: Fast segment anything. arXiv preprint arXiv:2306.12156  (2023)

\end{thebibliography}
\end{document}